%% file: main.tex
\documentclass[letterpaper,11pt,normalheadings,twocolumn,twoside=true]{article}

\usepackage[utf8]{inputenc}
\usepackage{lipsum}
\usepackage{fancyhdr}
\usepackage{cuted}
\usepackage{DejaVuSansCondensed}
\usepackage[condensed]{roboto}
\usepackage[sc]{mathpazo} 
\usepackage[T1]{fontenc}
\usepackage[babel,german=guillemets]{csquotes}
\usepackage[most]{tcolorbox}
\usepackage{titlesec}
\usepackage{csquotes}
\usepackage{enumitem}
\usepackage{xspace}
\usepackage{hyperref}
\usepackage[numbers,square]{natbib}
\usepackage{wrapfig}

\usepackage[margin=0.8in]{geometry}

\input{config}

\begin{document}

\input{title}

\maketitle

\begin{strip}
    \includegraphics[width=\textwidth]{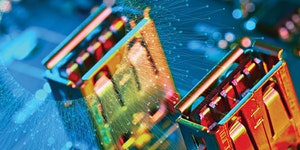}
\end{strip}

\input{introduction}

\input{current_state}

\input{program_synthesis}

\input{compiler_technology}

\footnotesize{
\bibliography{ref}{}
\bibliographystyle{plainnat}
}

\end{document}

%% file: config.tex
\newcommand{\headerfont}{
	\fontfamily{Roboto-TLF}\fontseries{b}\fontsize{7}{11}\selectfont}
\newcommand{\secfont}{%
	\fontfamily{Roboto-TLF}\fontseries{b}\fontsize{12}{11}\selectfont}
\newcommand{\subsecfont}{%
	\fontfamily{Roboto-TLF}\fontseries{b}\fontsize{11}{11}\selectfont}

\titlespacing{\section}{0pt}{15pt}{2pt}
\titlespacing{\subsection}{0pt}{6pt}{0pt}
\titleformat{\section}{\secfont}{}{0pt}{\MakeUppercase}
\titleformat{\subsection}{\subsecfont}{}{0pt}{}

\definecolor{light-blue}{HTML}{BED8F0}

\newcommand{\sidebar}[1]{
\begin{center}
\begin{tcolorbox}[notitle, sharp corners, colframe=light-blue,
colback=light-blue,boxrule=1pt, boxsep=0pt, enhanced,width=8cm]
\large \sffamily
#1
\end{tcolorbox}
\end{center}
}

%% file: title.tex
\title{}
\author{}
\date{}

\begin{titlepage}
        \vspace*{3in}
		\centering{
			{\fontsize{26}{40}\selectfont 
			Report of the Workshop on Program Synthesis for Scientific Computing}
		}\\
		\vspace{5mm}
		\centering{
			{\fontsize{16}{40}\selectfont August 4--5, 2020}
		}\\
		\vspace{10mm}
		\centering{\large{Organizers:}}\\
		\centering{\normalsize{Hal Finkel, Argonne National Laboratory}}\\
		\centering{\normalsize{Ignacio Laguna, Lawrence Livermore National Laboratory}}\\
		
		\vspace{10mm}

\newpage
\centering{\large{Participants:}}\\
\vspace*{5mm}
\scalebox{0.7}{
\begin{tabular}{lll}
Maaz Ahmad (University of Washington) & Zhengchun Liu (Argonne National Laboratory) \\
Alex Aiken (Stanford University) & Alexey Loginov (GrammaTech,  Inc.) \\
Aws Albarghouthi (University of Wisconsin - Madison) & Nuno Lopes (Microsoft Corporation) \\
Farhana Aleen (NVIDIA Corporation) & Abid Malik (Brookhaven National Laboratory) \\
Francis Alexander (Brookhaven National Laboratory) & Ruben Martins ( Carnegie Mellon University) \\
Rajeev Alur (University of Pennsylvania) & James McDonald (Kestrel Institute) \\
Saman Amarasinghe (Massachusetts Institute of Technology) & Thirimadura Mendis (Massachusetts Institute of Technology) \\
Todd Anderson (Intel Corporation) & Abdullah Muzahid (Texas A\&M University) \\
Pavan Balaji (Argonne National Laboratory) & Mayur Naik (University of Pennsylvania) \\
Prasanna Balaprakash (Argonne National Laboratory) & Sri Hari Krishna Narayanan (Argonne National Laboratory) \\
Ras Bodik (University of Washington) & Brandon Neth ( University of Arizona) \\
James Bornholt (University of Texas at Austin) & Matthew Norman (Oak Ridge National Laboratory) \\
Bill Carlson (Center for Computing Sciences) & Boyana Norris (University of Oregon) \\
Barbara Chapman (BNL \& Stony Brook University) & Cathie Olschanowsky (Boise State Univeristy) \\
François Charton (Facebook) & Cyrus Omar (University of Michigan) \\
Swarat Chaudhuri (University of Texas at Austin) & Peter-Michael Osera (Grinnell College) \\
Estee Chen (University of Pennsylvania) & Pavel Panchekha (University of Utah) \\
Alvin Cheung (University of California,  Berkeley) & Sheena Panthaplackel (University of Texas at Austin) \\
Amazon \& University of Chicago (spertus.com) & Eun Jung Park  (EJ) (Los Alamos National Laboratory) \\
Taylor Childers (Argonne National Laboratory) & Gilchan Park (Brookhaven National Laboratory) \\
Ravi Chugh (University of Chicago) & Madhusudan Parthasarathy (UIUC) \\
Valentin Clement (Oak Ridge National Laboratory) & Tharindu Patabandi (University of Utah) \\
Loris D'Antoni (University of Wisconsin - Madison) & Hila Peleg (University of California,  San Diego) \\
Arnab Das (University of Utah) & Ruzica Piskac (Yale University) \\
Isil Dillig (University of Texas at Austin) & Nadia Polikarpova (University of California,  San Diego) \\
Johannes Doerfert (Argonne National Laboratory) & Tobi Popoola (Boise State Univeristy) \\
Krzysztof Drewniak (University of Washington) & Xiaokang Qiu (Purdue University) \\
Anshu Dubey (Argonne National Laboratory) & Arjun Radhakrishna (Microsoft Corporation) \\
Souradeep Dutta (University of Pennsylvania) & Jonathan Ragan-Kelley (Massachusetts Institute of Technology) \\
Markus Eisenbach (Oak Ridge National Laboratory) & Krishnan Raghavan (Argonne National Laboratory) \\
Nur Aiman Fadel (Swiss National Supercomputing Centre) & Yihui Ren (Brookhaven National Laboratory) \\
Kyle Felker (Argonne National Laboratory) & Thomas Reps (University of Wisconsin - Madison) \\
Ian Foster (Argonne National Laboratory) & Kamil Rocki (Cerebras Systems) \\
Franz Franchetti (Carnegie Mellon University) & Jose Rodriguez (Intel Corporation) \\
Milos Gligoric (University of Texas at Austin) & Tiark Rompf (Purdue University) \\
Cindy Rubio Gonzalez (University of California, Davis) & Randi Rost (Intel Corporation) \\
Ganesh Gopalakrishnan (University of Utah) & Roopsha Samanta (Purdue University) \\
Justin Gottschlich (Intel Labs \& University of Pennsylvania) & Mark Santolucito (Barnard College) \\
Vinod Grover (NVIDIA Corporation) & Vivek Sarkar (Georgia Institute of Technology) \\
Mary Hall (University of Utah) & Markus Schordan (Lawrence Livermore National Laboratory) \\
Niranjan Hasabnis (Intel Corporation) & Eric Schulte (GrammaTech,  Inc.) \\
Amaury Hayat (Rutgers University) & Koushik Sen (University of California,  Berkeley) \\
Thomas Helmuth (Hamilton College) & Srinivasan Sengamedu (Amazon) \\
Michael Heroux (Sandia National Laboratories) & Dolores Shaffer (Science and Technology Associates, Inc.) \\
Paul Hovland (Argonne National Laboratory) & Min Si (Argonne National Laboratory) \\
Justin Hsu (University of Wisconsin - Madison) & Douglas Smith (Kestrel Institute) \\
Jan Hueckelheim (Argonne National Laboratory) & Armando Solar-Lezama (Massachusetts Institute of Technology) \\
Roshni Iyer (University of California, Los Angeles) & George Stelle (Los Alamos National Laboratory) \\
John Jacobson (University of Utah) & Michelle Strout (University of Arizona) \\
Xiao-Yong Jin (Argonne National Laboratory) & Yizhou Sun (University of California,  Los Angeles) \\
Beau Johnston (Oak Ridge National Laboratory) & Joseph Tarango (Intel Corporation) \\
Vinu Joseph (University of Utah) & Zachary Tatlock (University of Washington) \\
Ian Karlin (Lawrence Livermore National Laboratory) & Valerie Taylor (Argonne National Laboratory) \\
Vineeth Kashyap (GrammaTech, Inc.) & Supun Tennakoon (Purdue University) \\
Samuel Kaufman (University of Washington) & Aditya Thakur (University of California, Davis) \\
Sarfraz Khurshid (University of Texas at Austin) & Rajeev Thakur (Argonne National Laboratory) \\
Jungwon Kim (Oak Ridge National Laboratory) & Jesmin Jahan Tithi (Intel Corporation) \\
Martin Kong (University of Oklahoma) & Jeffrey Vetter (Oak Ridge National Laboratory) \\
Jaehoon Koo (Northwestern University) & Brice Videau (Argonne National Laboratory) \\
Siddharth Krishna (Microsoft Corporation) & Ashwin Vijayakumar (Intel Corporation) \\
Michael Kruse (Argonne National Laboratory) & Fei Wang (Purdue University) \\
Ignacio Laguna (Lawrence Livermore National Laboratory) & Wei Wang (University of California,  Los Angeles) \\
Jacob Lambert (University of Oregon) & Xinyu Wang (University of Michigan) \\
Edward Lee (University of California,  Berkeley) & Fangke Ye (Georgia Institute of Technology) \\
Seyong Lee (Oak Ridge National Laboratory) & Jisheng Zhao (Georgia Institute of Technology) \\
Richard Lethin (Reservoir Labs) & Shengtian Zhou (Intel Corporation) \\
Dmitry Liakh (Oak Ridge National Laboratory) & Tong Zhou (Georgia Institute of Technology) \\
Nevin Liber (Argonne National Laboratory) & \\
\end{tabular}
}
		
		\vspace*{\stretch{1}}
		\newpage
		\centering{\footnotesize This report
was prepared as an account of a workshop sponsored by the U.S. Department of Energy. Neither the United States Government nor any agency thereof, nor any of their employees or officers, makes any warranty, express or implied, or assumes any legal liability or responsibility for the accuracy, completeness, or usefulness of any information, apparatus, product, or process disclosed, or represents that its use would not infringe privately owned rights. Reference herein to any specific commercial product, process, or service by trade name, trademark, manufacturer, or otherwise, does not necessarily constitute or imply its endorsement, recommendation, or favoring by the United States Government or any agency thereof. The views and opinions of document authors expressed herein do not necessarily state or reflect those of the United States Government or any agency thereof. Copyrights to portions of this report (including graphics) are reserved by original copyright holders or their assignees and are used by the Government’s license and by permission. Requests to use any images must be made to the provider identified in the image credits (if any) or the first author.

The Workshop on Program Synthesis for Scientific Computing was held virtually on August 4--5 2020.
}\\

\end{titlepage}


%% file: introduction.tex
\section{Introduction}
Computational methods have become a cornerstone of scientific progress, with nearly every scientific discipline relying on computers for data collection, data analysis, and simulation. As a result, significant opportunities exist to accelerate scientific discovery by accelerating the development and execution of scientific software.

The first priority research direction of the U.S. Department of Energy's (DOE's) report on extreme heterogeneity~\cite{vetter2018extreme} and the ninth section of the DOE community's \textit{AI for Science} report~\cite{stevens2020ai}, among other sources, highlight the need for research in AI-driven methods for creating scientific software. This workshop expands on these reports by exploring how \emph{program synthesis}, an approach to automatically generating software programs based on some user intent~\cite{gottschlich:2018:mapl, gulwani:2017:program}---along with other high-level, AI-integrated programming methods---can be applied to scientific applications in order to accelerate scientific discovery.

We believe that the promise of program synthesis (also referred to as \emph{machine programming}~\cite{gottschlich:2018:mapl}) for the scientific programming domain is at least the following:
\begin{itemize}

    \item Significantly reducing the temporal overhead  associated with software development. We anticipate increases in productivity of scientific programming by orders of magnitude by making all parts of the software life cycle more efficient, including reducing the time spent tracking down software quality defects, such as those concerned with correctness, performance, security, and portability~\cite{alam:2019:neurips, alam:2016:isca, hasabnis:2020:controlflag}

    \item Enabling scientists, engineers, technicians, and students to produce and maintain high-quality software as part of their problem-solving process without requiring specialized software development skills~\cite{ragan-kelley:2013:pldi}
\end{itemize}

Program synthesis is an active research field in academia, national labs, and industry. Yet, work directly applicable to scientific computing, while having some impressive successes, has been limited. This report reviews the relevant areas of program synthesis work, discusses successes to date, and outlines opportunities for future work.

\subsection{Background on Program Synthesis}
Program synthesis represents a wide array of machine programming~\cite{gottschlich:2018:mapl} techniques that can greatly enhance programmer productivity and software quality characteristics, such as program correctness, performance, and security. Specifically, program synthesis incorporates techniques whereby the following may occur:
\begin{itemize}
    \item The desired program behavior is specified, but the (complete) implementation is not. The synthesis tool determines how to produce an executable implementation.
    \item The desired program behavior or any partial implementation is ambiguously specified. Iteration with the programmer, data-driven techniques, or both are used to construct a likely correct solution.
\end{itemize}

\begin{figure}
\includegraphics[width=\linewidth]{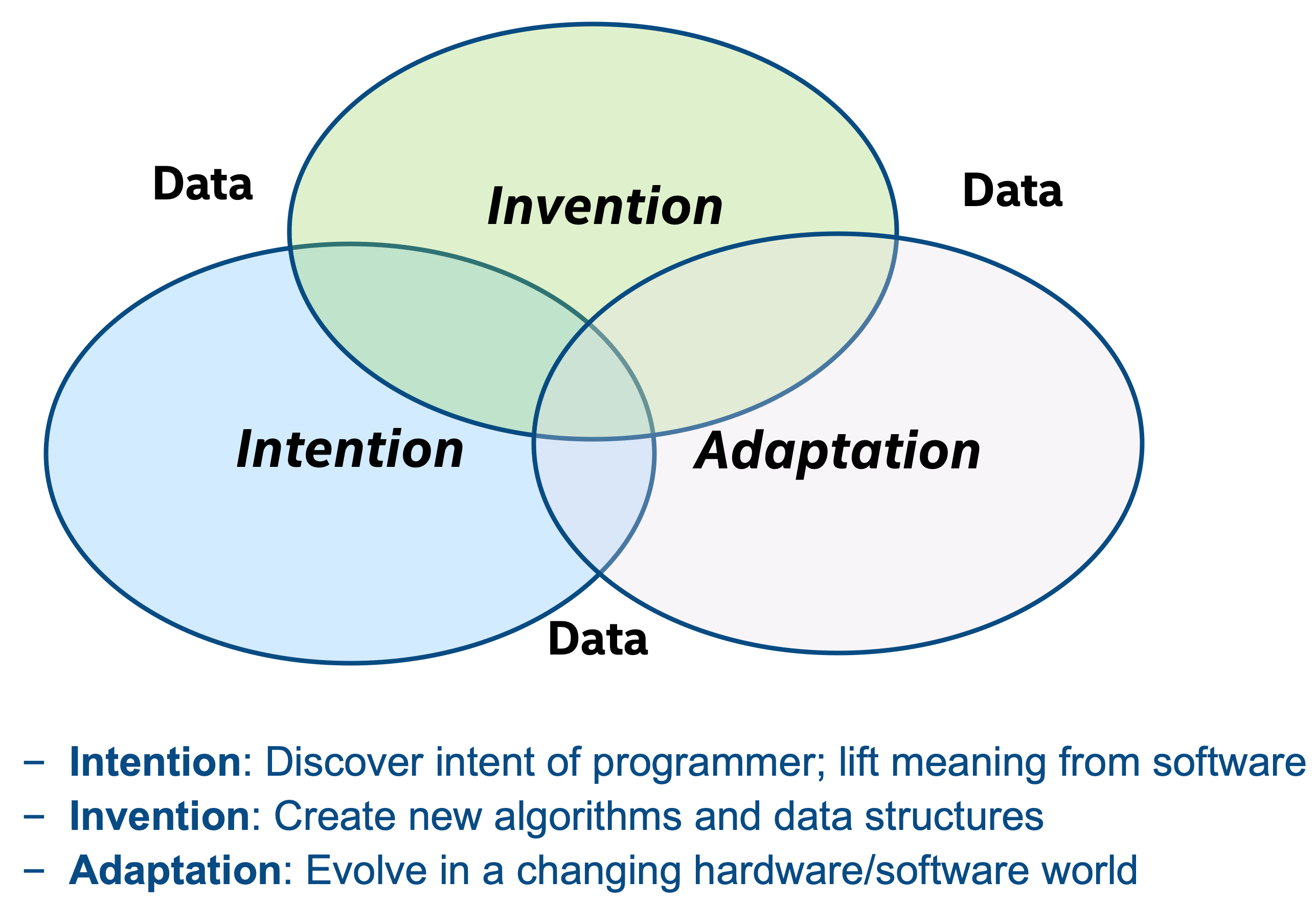}
\caption{The Three Pillars of Machine Programming (credit: Gottschlich et al.~\cite{gottschlich:2018:mapl}).}\label{fig:three_pillars}
\end{figure}

In \emph{"The Three Pillars of Machine Programming"} nomenclature, program synthesis is classified in the space of \emph{intention}~\cite{gottschlich:2018:mapl} (see Figure~\ref{fig:three_pillars}). The goal of intention is to provide programmers (and non-programmers) with ways to communicate their ideas to the machine. Once these goals have been expressed, the machine  constructs (\emph{invents}) a higher-order representation of the user's intention. Next, the machine programming system is free to \emph{adapt} the higher-order invented software representation to a lower-order representation that is specific to the user's unique software and hardware environment. This process tends to be necessary to ensure robust software quality characteristics such as performance and security.

When the implementation---the \textit{how} of the program---is unambiguously specified, the process of translating that specified implementation into an executable program is called compilation. 

Good compiler technology is essential to scientific programming; and as discussed later in this report,  opportunities exist for enhancing compiler technology to better enable program synthesis technology. These two areas, program synthesis and compilers, inform each other; and as we look toward the future, state-of-the-art programming environments are likely to contain both synthesis and compilation capabilities.

\sidebar{Program synthesis is expected to reduce software development overhead and increase confidence in the correctness of programs. Compiler technology will be essential to enable this technology.}

The first program-synthesis systems focused on the automated translation of mathematically precise program specifications into executable code. These systems performed what is sometimes called \textit{deductive synthesis} and often functioned by trying to match parts of the specification to a library of relevant implementation techniques.

With the advent of strong satisfiability modulo theories (SMT) solvers, program synthesis had an important new technology on which to build. An SMT solver can naturally produce counterexamples to an inconsistent set of mathematical assertions or produce an assertion that no counterexamples exist, which is useful for several kinds of verification and synthesis tasks. Writing mathematical specification is often subtle, however, and considerable attention in the synthesis community has  focused on \textit{inductive synthesis}: the synthesis of programs based on behavioral examples~\cite{flashfill}.

Of course, systems can be both deductive and inductive, which is useful because sometimes a programmer knows some of the desired properties but wishes to fill in the remaining information necessary to construct the program using examples. 
For example, techniques for \emph{type-and-example-directed synthesis} deductively use types 
provided by the programmer to guide an inductive search for missing program fragments that 
satisfy the given examples \cite{DBLP:conf/pldi/OseraZ15,DBLP:journals/pacmpl/LubinCOC20}.
Combining inductive synthesis with SMT solver iteration, using solver-generated counterexamples to guide the synthesis process, has also been a fruitful area: counterexample-guided inductive synthesis (CEGIS) has produced exciting results over the past decade.

CEGIS and related techniques are good at refining potential solutions that are structurally close to a correct answer, but  the techniques have more difficulty in  searching the unbounded space of potential program structures. Evolutionary algorithms have made important contributions to this problem, especially those that encourage the selection of specialists (potential solutions that work for some, but not necessarily all, solution objectives). 

The deep learning revolution has led to significant advancements in program synthesis as well. The space of potential program structures can be explored by using differentiable programming, reinforcement learning, and other machine learning techniques. Deep learning has also made practical the incorporation of natural language processing into the synthesis process and the generation of natural language comments as part of the synthesis output. Machine learning techniques now power advanced autocomplete features in various development environments, and looking toward the future, advanced tools can explore more than single-statement completions.

As program-synthesis technology, driven by advanced deep learning, evolutionary, and verification techniques, moves toward tackling real-world programming problems, imparting the resulting productivity gains to scientific programming will require techniques and capabilities that might not be required for other domains. In this report, we  explore the state of and challenges in scientific programming and how research in program synthesis technology and synergistic research in compiler technology might be directed to apply to scientific-programming tasks.

%% file: current_state.tex
\section{Current State of Scientific Application Development}

Scientific application development is essential to scientific progress; and yet, while advances in both scientific techniques and programming tools continue to improve programmer productivity, creating state-of-the-art scientific programs remains challenging.

\begin{figure*}
\includegraphics[width=\linewidth]{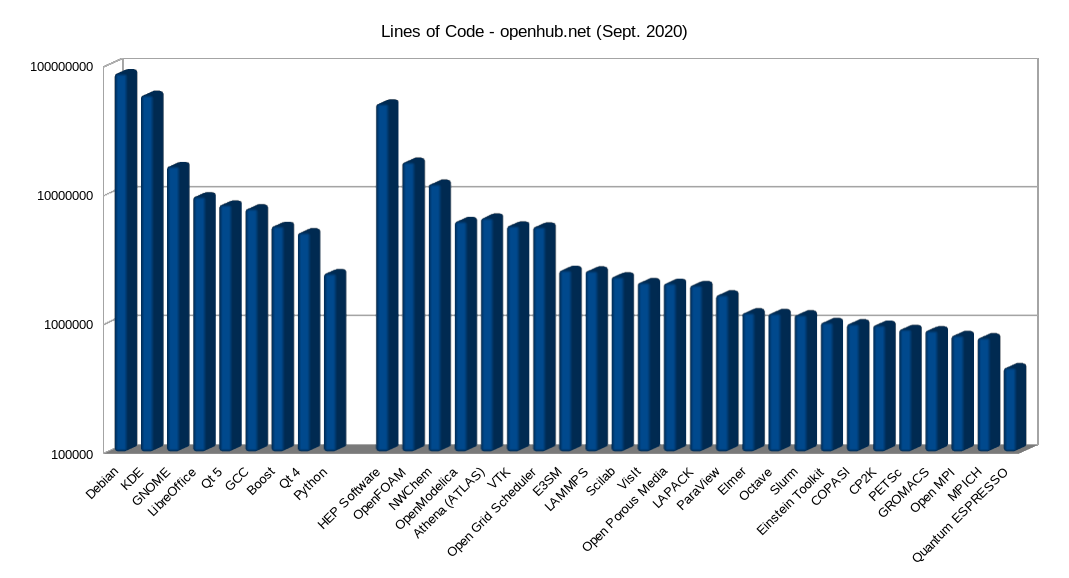}
\caption{Number of source lines of code in various packages: general packages on the left, scientific packages on the right. Nearly all data from openhub.net (Sept. 2020).}\label{fig:science-loc}
\end{figure*}

Software complexity has increased over the past decades at an astounding rate, with many popular applications and libraries containing tens of millions of lines of code. As shown in Figure~\ref{fig:science-loc}, scientific development has not been immune from this increase in complexity. 

General software infrastructures, along with algorithmic and mathematical techniques, have become increasingly sophisticated. Hardware architectures and the techniques necessary to exploit them have also become increasingly sophisticated, as has the science itself. Managing the resulting complexity is difficult even for the most experienced scientific programmers. Moreover, a lot of scientific programming is not done by experienced programmers but  by scientific-domain students and recent graduates with only a few years of experience~\cite{milewicz2019characterizing}. As a result,  new programmers on a project have difficulty reaching high levels of productivity.

\subsection{The Largest Challenges}

Code development itself is labor intensive; and in order to develop scientific applications, significant portions of the development require direct input from domain experts. These applications often contain critical components that are mathematically complex; and hence developers require advanced mathematical abilities, strong programming skills, and a good understanding of the science problem being solved. 

Separations of concerns are common, and  not every scientific-software developer implements every mathematical technique from scratch. Nevertheless, developers  need sufficient knowledge of the relevant mathematical techniques to select and use applicable libraries.

\sidebar{Scientific software has become increasingly complex over the years. Tools and techniques to improve productivity in this space are desperately needed.}

Moreover,  often  the complete design for the application cannot be specified up front.  One may not  know ahead of time what grid resolutions, discretization techniques, solvers, and so on will work best for the target problems. Instead, development is an iterative process that is part of the scientific investigatory process. As the science evolves, the target problem might change as well, necessitating significant changes to the application design.

While application development for dynamic consumer markets also faces challenges with evolving requirements, scientific software must often meet tight performance and mathematical requirements, where large changes in design must be implemented quickly by small development teams, imparting unique needs for productivity-enhancing tools in this space.

\textbf{Adapting to New Hardware:} Scientific computing applications tend to require high computational performance and, as a result, are structured and tuned to maximize achieved performance on platforms of interest. However, computational performance is maximized on cutting-edge hardware, and cutting-edge hardware has been evolving rapidly. As a result, applications have had to adapt to new CPU features, such as SIMD vector registers, GPU accelerators, and distribution over tens of thousands of independent nodes. These architectures continue to evolve, with corresponding changes to their programming models, and applications must adapt in turn. If significant work is invested in turning for a particular architecture, as is often the case~\cite{aleen:2016:cf}, repeating that level of work for many different kinds of systems is likely infeasible.

To maintain developer productivity in the face of a variety of target architectures, the community has placed  significant focus on programming environments that provide some level of \textit{performance portability}. The idea is that reasonable performance, relative to the underlying system capabilities for each system, can be obtained with no, or minimal, source code changes between systems. 

What qualifies as reasonable performance and what qualifies as minimal changes to the source code are hotly debated and, in the end, depend on many factors specific to individual development teams. Nevertheless, overall goals are clear, and these motivate the creation of compiler extensions (e.g., OpenMP), C++ abstraction libraries (e.g., Kokkos~\cite{edwards2014kokkos}), and domain-specific languages (e.g., Halide~\cite{ragan2013halide}). 

Furthermore, the design of these portable abstractions is largely reactionary; and while one can anticipate and incorporate some new hardware features before the hardware is widely available, often the best practices for using a particular hardware architecture are  developed only after extensive experimentation on the hardware itself. This situation leads to a natural tension between the productivity gain from using portable programming models and the time-to-solution gain potentially available from application- and hardware-specific optimizations. 

The ability to perform autotuning on top of these technologies has been demonstrated to enhance performance significantly~\cite{Balaprakash:IEEE18}. Some systems (e.g., Halide) were specifically designed with this integration in mind. However, autotuning brings with it a separate set of challenges that make deployment difficult. If the autotuning process is part of the software build process, then the build process becomes slow and potentially nondeterministic. 

On the other hand, if the autotuning process produces artifacts that are separately stored in the source repository, then the artifacts need to be kept up to date with the primary source code, a requirement made more difficult by the fact that not all developers have access to all of the hardware on which the artifacts were generated. A potential solution to these challenges is to perform the autotuning while the application is running, but then the time spent autotuning must be traded against the potential performance benefits. Nevertheless, the optimal tuning results sometimes depend on the state of the application's data structures (e.g., matrix sizes), and these can change during the course of a long-running process, providing an additional advantage to during-execution autotuning and autotuning procedures that can make use of detailed profiling data.

High performance can often be obtained by using different implementations of the same algorithm, and the aforementioned frameworks naturally apply to this case. Sometimes, however,  especially when accelerators are involved, different algorithms are needed in order to obtain acceptable performance on different kinds of hardware. In recognition of this reality, a number of capabilities have been explored for supporting algorithmic variants that can be substituted in a modular fashion as part of the porting process (e.g., OpenMP metadirectives, PetaBricks~\cite{ansel2009petabricks}). Having multiple available algorithms for tasks within an application, however, makes testing and verification of the application more difficult. Development and maintenance are also more expensive because each algorithmic variant must be updated as the baseline set of required features expands over time and as defects are fixed.

\sidebar{The ability of scientific codes to quickly adapt to new hardware is increasingly challenging. While performance-portable programming models and autotuning help, more advanced end-to-end capabilities are needed to adapt algorithms and data structures to new environments.}

\textbf{Data Movement Cost:}
The performance of many scientific applications is dominated by data movement. Customizing temporary storage requirements to architectures and application needs has demonstrated promising performance improvements~\cite{olschanowsky2014study}. 
Previous work combined scheduling transformations within a sparse polyhedral framework and dataflow graphs to enable human-in-the-loop architecture-specific optimization~\cite{davis2018transforming}. Because of large variations in the design of memory subsystems, optimizing the actual layout of data in memory can reduce data movement as well as reduce its cost.  As one example, bricks---mini-subdomains in contiguous memory organized with adjacency lists---are used to represent stencil grids~\cite{Zhao:SC19,Zhao:PPoPP21}.  Bricks reduce the cost of data movement because they are contiguous and therefore decrease the number of pages and cache lines accessed during a stencil application.  Through a layout of bricks optimized for communication, we also eliminate the data movement cost of packing/unpacking data to send/receive messages.

\textbf{Scheduling:} A large fraction of scientific applications are still written in legacy code such C/C++ and require mapping to multithreaded code either manually or by autoparalellizing compilers. Performant execution of such legacy codes is dependent on good task scheduling.  While most scheduling techniques perform statically, Aleen et al.~\cite{aleen:2010:ppopp} showed that orchestrating dynamically (by running a lightweight emulator on the input-characterization graph extracted from the program) can better balance workloads and provide further improvement over static scheduling. 

\begin{figure}[h]
\includegraphics[width=\linewidth]{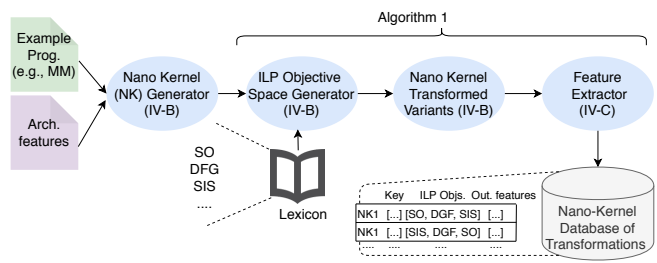}
\caption{\label{fig:chelini:pact2020:offline} Offline Database Construction}
\end{figure}

\begin{figure}[h]
\includegraphics[width=\linewidth]{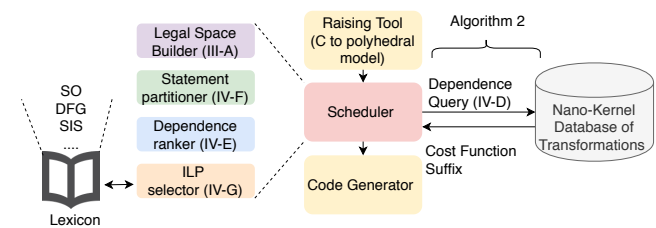}
\caption{\label{fig:chelini:pact2020:online} Adaptive Scheduling and Objective Selection (Database Querying).}
\end{figure}

{\bf Polyhedral Multiobjective Scheduling:}
As a step toward achieving portable performance,  Kong and Pouchet \cite{kong.pldi.2019} recently proposed a extensible kernel set of integer linear program (ILP) objectives that can be combined and reordered to produce different performance properties. Each ILP objective aims to maximize or minimize some property on the generated code, for instance, minimizing the stride penalty of the innermost loop.  However, this work did not address how to select the objectives to embed into the ILP. More recently, Chelini et al.~\cite{chelini2020automatic} proposed a systematic approach to traverse the space of ILP objectives previously defined by first creating  an offline database of transformations. Such a database is constructed from a number of input cases exhibiting different dependence patterns, which are then transformed via all the possible 3-permutations of ILP objectives from the kernel set of transformations. 
The resulting transformed codes are then analyzed to extract specific code features that are stored together with the input dependencies and the transformations used (see Fig.~\ref{fig:chelini:pact2020:offline}). Later, during the compilation phase, the database is queried to adaptively select ILP objectives to embed. This process is illustrated in Fig.~\ref{fig:chelini:pact2020:online}.

\textbf{Testing and Verification:} The correctness of scientific-computing applications is critical because the scientific enterprise depends on predictive computational techniques. Especially in areas where the results produced by scientific applications are used to ensure safety or otherwise inform public policy, incorrect results can cause  serious problems. In general, science and engineering are competitive fields, and erroneous results put those depending on them at a relative disadvantage. Across the board, productivity can be severely hampered by application crashes and the time spent diagnosing and fixing misbehaving code.

Writing tests for scientific applications is often difficult and time consuming. As in any other software project, individual components should be tested with unit tests, and in addition, application-level tests are essential. If tests represent a fixed set of known input-output pairs for each component or for the application, covering all of the various combinations of allowed features and behaviors generally requires an exponential number of test cases. For tests that, individually, automatically explore more of the allowed state space, verifying the correct behavior is difficult. Programmers often fall back on verifying only invariants, not the complete output itself. This approach is especially true for physical-simulation results where exact answers are not known. Invariants (where they exist) such as conservation of energy and the application not crashing might be the only things that the tests actually end up checking. Tests providing higher-confidence verification, including comparisons with competing applications, and numerical-convergence analysis are often not automated and hence are performed manually only on irregular occasions (e.g., just before a major release). Problems discovered during these manual tests, as with other problems discovered late in the development cycle, are often difficult to diagnose and correct. The following are some notable challenges.
\begin{itemize}
    \item \textbf{Numerics:} Many physical systems exhibit nontrivial sensitivity to their initial conditions, and thus small differences from truncation error, round-off error, and other small perturbations legitimately cause large differences in the final results. Verification in these cases is subtle, sometimes relying on statistical properties to compare with known solutions and sometimes relying on physical invariants; but even so, picking good thresholds for numerical comparisons is often done by manual experimentation and rules of thumb.
    \item \textbf{Asynchronous Execution: } Modern hardware demands the use of concurrent, and often parallel, execution in order to take advantage of the available computational capabilities. This makes testing and verification difficult because concurrent execution is nearly always nondeterministic. It is difficult to be sure that a particular algorithmic implementation is free from race conditions and other constraint violations under all possible execution scenarios: only a finite number, often a small part of the overall space of possibilities, are exhibited during testing. Pairing that testing with special programs designed to detect race conditions (e.g., valgrind, Thread Sanitizer) can help, but these programs add overheads and thus additional trade-offs into the overall testing process. Asynchronous execution frameworks (e.g., OpenMP tasks) generally depend on programmer-declared dependencies to ensure valid task scheduling, and mistakes in these dependency declarations are sometimes difficult to detect. 
    \item \textbf{Large Scale:} Scientific simulations often need to run at large scale in order to exhibit behaviors relevant for testing. Should something go wrong, debugging applications running in parallel on tens of thousands of nodes is difficult. Runs of an application, which are often dispatched from batch queues at unpredictable times, are expensive to recreate; and while interactive debugging sessions can be scheduled, turnaround times for large-scale reservations are not fast, and even state-of-the-art debugging tools for parallel applications have difficulty scaling to large systems. In some cases, state capture and mocking can be used to replicate and test relevant behaviors at smaller scales, but these are  time consuming to implement.
    \item \textbf{System Defects:} At leadership scale, systems and their software push the state of the art and, as a result, may themselves have defects that lead to incorrect application behaviors. For the largest machines, vendors are often unable to test their software (e.g., their MPI implementation) at the full scale of the machine until that machine is assembled in the customer's data center. Early users of these machines frequently spend a lot of time helping track down bugs in hardware, system software, and compilers. For these users, the benefit from being able to run unprecedented calculations provides the motivation to continue even in the face of these kinds of issues, but getting things working at large scale is often more difficult than one might imagine.
    \item \textbf{Performance:} It is often desirable for testing to cover not only correctness but also performance. Application performance in scientific computing is often a critical requirement because only a limited number of core-hours are generally available in order to carry out particular calculations of interest. Testing of performance is difficult, however, both because many testing systems run many processes in parallel, making small-scale performance measurements noisy, and because many performance properties can  be observed only with large problem sizes at larger scales. Special tests that extrapolate small-scale performance tests can sometimes be constructed, but these are often done manually because the automation would involve even more work.
    \item \textbf{Resource Management:} As scientific software becomes more complex, problems often appear due to unfortunate interactions between different components. A common issue is resource exhaustion, where resources such as processor cores, memory, accelerators, file handles, and disk space, which are adequately managed by the different components in isolation, are not managed well by the composition of the components. For example, different components often allocate system memory with no regard for the needs of other components. Dedicated tests for resource usage can be written, but in practice this kind of testing is also performed manually. Many of these issues  surface only when problems are being run at large scale.
\end{itemize}

\subsection{Requirements for Automation}

Increasing the amount of automation in the scientific application development process can increase programmer productivity, and program synthesis can play a key role. In order to be truly helpful, automated tools need to be integrated into the iterative and uncertain scientific discovery process. Experience has taught us that a number of important factors should be considered.
\begin{itemize}
    \item \textbf{Separation of Concerns: } Tools must be constructed and their input formats designed  recognizing that relevant expertise is spread across different users and different organizations. A domain scientist may understand very well the kinds of mathematical equations that need to be solved, an applied mathematician may understand very well what kinds of numerical algorithms best solve different kinds of equations, and a performance engineer may understand very well how to tune different kinds of algorithms for a particular kind of hardware; but these people might not work together directly. As a result, it must be possible to compartmentalize the relevant knowledge provided to the tool, to enable both reuse and independent development progress. To the extent that tools support providing mathematical proofs of correctness, the information needed for these proofs should be providable on component interfaces, so the modularity can improve  the efficiency, understandability, and stability of the proof process.
    
    \item \textbf{Testing, Verification, Debugging: } Tools require specific features in order to assist with verification and debugging. Any human-generated input can, and likely will, contain mistakes. Mistakes can  take the form of a mismatch between the programmer's intent~\cite{gottschlich:2018:mapl} and the provided input and can also stem from the programmer's overlooking some important aspect of the overall problem. Moreover, tools themselves can have defects. Thus,  tools must be constructed to maximize the ability of programmers to find their own mistakes and isolate tool defects~\cite{LeeIPDPS14}. To this end,  tools must generally produce information on what they did, and why, and  provide options to embed runtime diagnostics helpful for tracking down problems during execution. FPDetect~\cite{das2020efficient} is an example of a tool providing sophisticated, low-overhead diagnostics that might be integrated with an automation workflow. Another example of synthesizable error detectors that can trap soft errors as well as bugs associated with incorrect indexing transformations is FailAmp~\cite{briggs2020failamp}, which, in effect, makes errors more manifest for easier detection.
    
    \item \textbf{Community and Support: } State-of-the-art science applications are complex pieces of software, often actively used for decades, and some are developed by large communities. Tools automating this development process need to integrate with existing code bases and support the complexity of production software. Actively used tools need to be supported by an active team. The productivity loss from issues with an undersupported tool, whether defects or missing features, can easily overwhelm the gains from the tool itself. Moreover, tools that generate hardware-specific code need continual development to support the latest hardware platforms. All tools need to evolve over time to incorporate updated scientific and mathematical techniques. Support and adoption are helped by using tools that integrate with popular languages and programming environments, such as C++ and C++ libraries (e.g., Kokkos). Tools making use of production-quality compiler components (e.g., Clang as a library) tend to be best positioned for success. Sometimes, however, adoption of such tools is hampered by existing code bases in older or less widely used languages (e.g., C, Fortran), and an opportunity exists for automation tools to assist with translating that code to newer languages.
\end{itemize}

%% file: program_synthesis.tex
\section{Program Synthesis}

\subsection{Existing Work in Scientific Programming}

In scientific computing, domain-specific languages have a long history, and scientists and engineers have increased the productivity of programming by creating specialized translators from mathematical expressions to code in C, C++, or Fortran. While many of these tools have an implicit implementation specification, they are instructive in the synthesis context as demonstrations of the kinds of high-level descriptions of mathematical calculations that programmers find useful. Claw~\cite{clement2018claw}/GridTools~\cite{osuna2019report} (stencil generation for climate modeling), Kranc~\cite{husa2006kranc} (stencil generation for numerical relativity), FireDrake~\cite{rathgeber2016firedrake}/FEniCS~\cite{logg2012ffc} (for the generation of finite-element solvers for partial-differential equations), lbmpy~\cite{bauer2020lbmpy} (for the generation of lattice Boltzmann simulations), TCE~\cite{hirata2003tensor} (for the generation of tensor-contraction expressions), and many others have demonstrated the utility of code generation from physical equations. Likewise, for traditional mathematical kernels, high-level code generators have been produced (e.g., SPIRAL~\cite{franchetti2018spiral}, LGen~\cite{spampinato2014basic}, Linnea~\cite{barthels2020automatic}, TACO~\cite{kjolstad2017tensor}, Devito~\cite{lange2016devito}), and the codelet generator in FFTW~\cite{frigo1998fftw}). Some of these tools, such as SPIRAL, perform synthesis as well as code generation because they discover new algorithms from the exploration of relevant mathematical properties.

\begin{figure*}
\includegraphics[width=\linewidth]{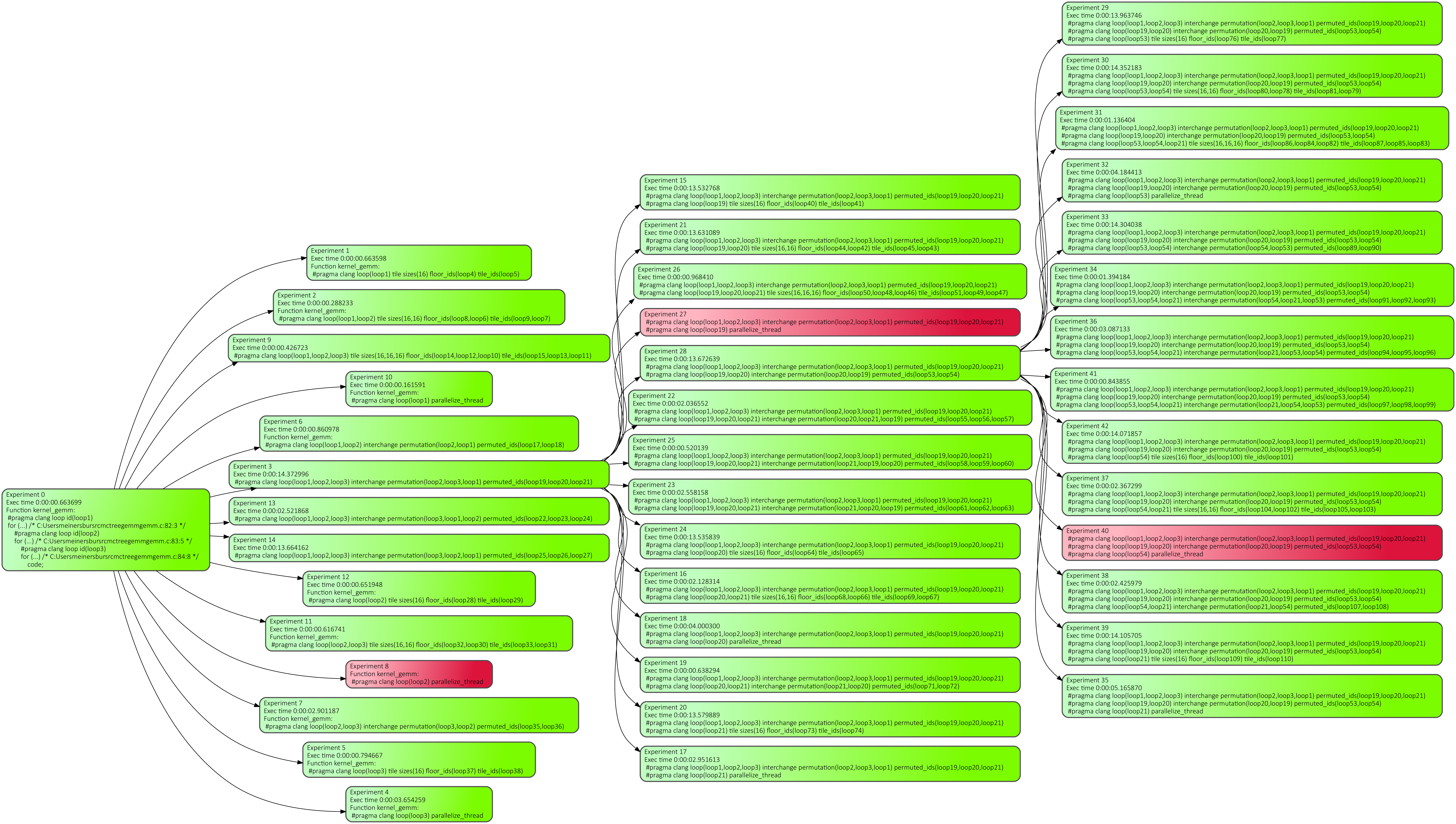}
\caption{Autotuning search space for composed loop transformations}\label{fig:looptransform-tree}
\end{figure*}

Autotuning is a common technique used to produce high-performance code, and machine learning can be used to dynamically construct surrogate performance models to speed up the search process. Autotuning can be used simply to choose predetermined parameter values or to explore complex implementation spaces and anything between. Stencil code generators such as PATUS~\cite{christen11-patus} and LIFT~\cite{steuwer16-lift} use autotuning to find the best tile sizes. ATLAS~\cite{whaley2001automated} uses this technique to select implementation parameters for linear algebra kernels. Recent work, making use of the ytopt~\cite{ytopt} autotuning framework, has explored directly searching complex hierarchical spaces of loop nest transformations, as shown in Figure~\ref{fig:looptransform-tree}, using a tree search algorithm~\cite{llvmhpc20-mctree}. Some domain-specific languages, such as Halide, have been designed with a separate scheduling language that can be used by this kind of autotuning search technique.

To synthesize code that examines runtime data to perform optimizations, 
inspector/executor strategies~\cite{Mirchandaney88,Saltz91} employ inspector code that at runtime determines data reorderings, communication schedules for parallelism, and computation reorderings that are then used by the transformed executor code~\cite{Saltz97,Stichnoth97,DingKen99,MitchellCarFer99,Mellor-Crummey2001,HanT:TPDS06,Wu2013,Basumallik06}.  Strout and others have developed inspector/executor strategies such as full sparse tiling that enable better scaling on the node because of reduced memory bandwidth demands~\cite{StroutLCPC2002,Strout14IPDPS,Demmel08,Ravishankar12SC}.  Ravishankar et al.~\cite{Ravishankar2015} composed their distributed-memory parallelization inspector/executor transformation with affine transformations that enable vector optimization.  Compiler support for such techniques consists of program analyses to determine where such parallelization is possible and the compile-time insertion of calls to runtime library routines for performing aspects of the inspector~\cite{rauchwerger95scalable}. 

Some early work applying general-purpose synthesis techniques to scientific computing problems has appeared. For example, AccSynt~\cite{collieprogram} applied enumerative synthesis to the problem of generating efficient code for GPUs. AutoPandas~\cite{bavishi2019autopandas} uses machine-learning-based search to automate the programming of data-table-manipulating programs. Lifting, the process of extracting a high-level model from the source code of an existing implementation, is important for handling legacy code and extracting knowledge from existing code bases~\cite{DBLP:conf/pldi/KamilCIS16, DBLP:conf/pldi/CheungSM13, DBLP:conf/sigmod/AhmadC18}. Dexter~\cite{ahmad2019automatically}, for example, can translate C++ kernel implementations into Halide, and OpenARC~\cite{ACC2FPGA-ICS18} can automatically translate existing OpenACC/OpenMP programs to OpenCL specialized for FPGAs.

\subsection{Existing Work in Other Areas}

A lot of existing work on program synthesis has focused on general programming tasks, such as  writing functions that operate on common data structures like strings, lists, and trees  \cite{DBLP:conf/pldi/OseraZ15,DBLP:journals/pacmpl/LubinCOC20}. Recent work has tended to focus on tools that use natural language, behavioral examples, or both as input. The most widely deployed synthesis system is perhaps the Flash Fill feature in Microsoft Excel 2013+. Flash Fill can synthesize a small program to generate some column of output data given some columns of input data and some filled-in examples. The combination of deep learning with the large amount of publicly available source code (e.g., on GitHub) has provided opportunities for machine learning methods to capture knowledge from existing code bases at a massive scale. The resulting models can then be used for relatively simple tasks, such as autocomplete capabilities in editors, but can also drive complex search techniques in sophisticated synthesis systems. Past academic work in restricted domains, such as SQLizer~\cite{yaghmazadeh2017sqlizer}, which translates natural languages to SQL queries, has begun inspiring commercial implementations. One interesting aspect of mining data from version-control repositories is that not only is the code itself available but its development history can also provide critical data. Systems that learn from past commits have been able to learn likely defects and can suggest fixes. Sketch~\cite{solar2007sketching}, Rosette~\cite{torlak2013growing}, Bayou~\cite{murali2017neural}, Trinity~\cite{martins_trinity_2019}, HPAT~\cite{totoni2017hpat}, and many others have explored different aspects of data-driven synthesis.

The autocomplete features in modern development environments are starting to incorporate advanced machine learning technology. Microsoft, for example, has built complex programming-language models for use with its technologies, and both Codota and Tabnine provide intelligent completion for several programming languages. Commercial tools have started to offer more advanced programming assistance, such as suggesting code refactorings to increase the productivity of maintenance tasks. Expanding on these techniques, researchers have demonstrated unsupervised translation between different programming languages~\cite{lachaux2020unsupervised}. In addition, considerable effort has been put into developing techniques for program repair. In the future, these kinds of technologies can be made available for scientific programmers as well, with models customized for relevant programming languages and libraries.

\sidebar{Integrating sophisticated example- and type-driven program synthesis and automated program-repair techniques into modern development environments remains an ongoing challenge.}

\emph{IDE Integration.}
Integrating sophisticated example- and type-driven program synthesis and automated program-repair techniques into modern development environments remains an ongoing challenge.  The Mnemosyne project\footnote{\url{https://grammatech.gitlab.io/Mnemosyne/docs/}} provides a framework for the integration of program synthesis techniques for code generation (e.g., Trinity~\cite{martins_trinity_2019}), as well as type inference~\cite{pradel2019typewriter,wei2020lambdanet} and test generation.\footnote{\url{https://hypothesis.readthedocs.io/en/latest/}} In this system independent synthesis \emph{modules} communicate with each other and with a user's IDE using Microsoft's Language Server Protocol.\footnote{\url{https://microsoft.github.io/language-server-protocol/}}  Programmatic communication between modules enables workflows in which multiple modules collaborate in multiphase synthesis processes. For example, the results of an automated test generation process may trigger and serve as input to subsequent automated code synthesis or program repair processes.

Another significant challenge is that program synthesis is often requested when the 
program is \emph{incomplete}, thst is, when there are missing pieces or errors that the programmer
hopes the synthesizer can help with. In other situations, however, standard
techniques for parsing, typechecking, and evaluation fail. Some or all of these may be necessary for the synthesizer to proceed. Recent work on formal reasoning about incomplete programs by using \emph{holes} has started to address this issue \cite{DBLP:conf/popl/OmarVHAH17,DBLP:journals/pacmpl/OmarVCH19}, and the Hazel programming environment is being designed specifically around this hole-driven development methodology \cite{DBLP:conf/snapl/OmarVHSGAH17}. 
Programs with holes are also known as program \emph{sketches} in the program synthesis community \cite{DBLP:conf/aplas/Solar-Lezama09}. 
Recent work on type-and-example directed program sketching that takes advantage of these modern advances in reasoning about incomplete programs 
represents a promising future direction for human-in-the-loop program synthesis \cite{DBLP:journals/pacmpl/LubinCOC20}.

\emph{Reversible Computation.} The reversible computation paradigm extends the
traditional forward-only mode of computation with the ability to
compute deterministically in both directions, forward and backward. It allows
the program to reverse the effects of the forward execution and go backwards to a
previous execution state.
\emph{Reverse execution} is based on the idea that for many
programs there exists an inverse program that can uncompute all
results of the (forward) computed program. The inverse program can be
obtained automatically either by generating reverse code from
a given forward code or by implementing the program in a reversible
programming language; and its compiler offers the capability to automatically generate both the forward and the inverse program. Alternatively, an interpreter for a reversible language can execute a program in both directions, forward and backward. For example, the reverse C compiler presented in \cite{perumalla2013}
generates reverse C code for a given C (forward) code. The imperative
reversible language Janus~\cite{YokoyamaGlueck:2007:Janus} allows
both the interpretation of a Janus program, where every language construct
has standard and inverse semantics, and the generation of forward and
reverse C code for a given Janus program.\footnote{Online Janus
interpreter at \url{https://topps.diku.dk/pirc/?id=janus\,.}}

Over the years, a number of theoretical aspects of reversible
computing have been studied, dealing with categorical foundations of
reversibility, foundations of programming languages, and term
rewriting, considering various models of sequential computations
(automata, Turing machines) as well as concurrent computations
(cellular automata, process calculi, Petri nets, and membrane
computing). An overview of the state of the art and use cases can be
found in \cite{Ulidowski2020ReversibleCE}, titled ``Reversible
Computation -- Extending Horizons of Computing.'' Reversible
computation has attracted interest for multiple applications, covering
areas as different as low-power computing \cite{landauer61},
high-performance computing with optimistic parallel discrete event
simulation \cite{schordan15,cingolani17,DBLP:journals/ngc/SchordanOJB18},
robotics \cite{laursen15}, and reversible debugging \cite{chen01}. In \cite{Schordan2020}, the generation of forward and
reverse C++ code from Janus code, as well as automatically generated
code based on incremental state saving, is systematically
evaluated. The example discussed in detail is a reversible variant of
matrix multiplication and its use in a benchmark for optimistic parallel discrete event simulation.

\emph{Automated Machine Learning.}  
Typically, the ML pipeline comprises several components: preprocessing, data balancing, feature engineering, model development, hyperparameter search, and model selection. Each of these components can have multiple algorithmic choices, and each of these algorithms can have different hyperparameter configurations. Configuring the whole pipeline is beyond human experts, who therefore often resort to trial-and-error methods, which are nonrobust and computationally expensive.  
Automated machine learning (AutoML) \cite{automl_book} is a technique for automating  the design and development of an ML pipeline. Several approaches developed for AutoML can be leveraged for program synthesis and autotuning. A related field is programming by optimization \cite{hoos2012programming}, where the algorithm designer develops templates and an optimization algorithm is used to compose these templates to find the right algorithm to solve a given problem. These methods have been used to develop stochastic local search methods for solving difficult combinatorial optimization problems.

\subsection{Expanding on Existing Work}

Existing work on program synthesis, while continuing to evolve to address programming challenges in a variety of domains, can be expanded  to specifically address the needs of the scientific programming community.

\begin{itemize}

    \item \textbf{Semantic Code Equivalence/Similarity: } The domain of semantic code similarity~\cite{Perry.SemCluster.PLDI19} aims to identify whether two or more code snippets (e.g., functions, classes, whole programs) are attempting to achieve the same goal, even if the way they go about achieving that goal (i.e., the implementation) is largely divergent. We believe this is one of the most critical areas of advancement in the space of machine programming. The reason is that once semantic code similarity systems demonstrate a reliable level of efficacy, they will likely become the backbone as well as enabling a deeper exploration of many auxiliary MP systems (e.g., automatic bug defection, automatic optimizations, repair)~\cite{ben-nun:2018:neurips, luan:2019:aroma, ye:2020;misim, Retreet,Dantoni.Qlose.CAV16,Perry.SemCluster.PLDI19}.

\sidebar{Program synthesis might address challenges inherent in targeting heterogeneous hardware architectures and generating performance-portable code.}

    \item \textbf{Hardware Heterogeneity: } Program synthesis might address challenges inherent in targeting heterogeneous hardware architectures, in cases where specialized code is needed (perhaps ``superoptimizations''),  where specialized algorithms are needed, and  where specialized ``scheduling'' functions are needed to dynamically direct work and data to the most appropriate hardware. Different data structures, in addition to different loop structures, may be required to achieve acceptable performance on different kinds of hardware.
    
    \item \textbf{Performance Portability: } Program synthesis might address the challenge of generating code that performs well across a wide variety of relevant hardware architectures~\cite{Sabne:MICRO15}. This synthesis procedure might account for later abilities to autotune the implementation for different target architectures.
    
   \item \textbf{Data Representation Synthesis: }
   Optimizing data movement for performance portability demands the ability to synthesize the representation of data to take advantage of hardware features and 
   input data characteristics such as sparsity.  Data representation synthesis includes data layout considerations and potentially should be coordinated with storage mappings that specify storage reuse.  Complex interactions among algorithms, memory subsystems, available parallelism, and data motivate delaying data structure selection until adequate information is available to make beneficial decisions.  Existing data structure synthesis research~\cite{Hawkins11,Loncaric2018,DBLP:journals/pvldb/YanC19} has focused on more general relational or map-based data structures and must be extended to scientific computing domains.

    \item \textbf{Numerical Properties: } Scientific programs are often characterized by numerical requirements for accuracy and sensitivity. Synthesis techniques might address the challenges in finding concrete implementations of mathematical algorithms meeting these requirements. This problem is made more difficult because these requirements are often not explicitly specified. Instead, requirements might  be only indirectly specified by the need for some postprocessing step to meet its requirements. In addition, the properties being extracted from the output of the algorithm might be fundamentally statistical in nature (e.g., a two-point correlation function), making verification of the program properties itself a fundamentally statistical process.

    \item \textbf{Workflows: } Program synthesis might address the challenges in correctly and efficiently composing different analysis and simulation procedures into an overall scientific workflow. This work might include the generation of specialized interfaces to enable coupling otherwise-modular components with low overhead. It might also include helping  use existing APIs in order to combine existing components to accomplish new tasks.
    
    \item \textbf{Translation: } The existing code base of scientific software contains large amounts of code in C, C++, Fortran, Python, and other languages. Much of this code lacks high-level descriptions;  moreover, reusing codes together that are written in different programming languages is often difficult. Worse, even within the same language, different parallel or concurrent programming models do not compose, for example, Kokkos vs.\ OpenMP vs.\ SYCL in C++, OpenMP vs.\ OpenACC in C/Fortran~\cite{SC20:CCAMP}, Dask vs.\ Ray vs.\ Parsl~\cite{babuji2019parsl} in Python. Program synthesis and semantic representations, such as Iyer et al.'s program-derived semantics graph~\cite{iyer:2020:PSG}, can help by performing lifting and translation of existing code between different languages and representations and by providing better documentation.
\end{itemize}

Program synthesis can act as an automated facility; but given the often ambiguous nature of expressed programming goals, synthesis tools are expected to interact with programmers in a more iterative manner. A program synthesis tool can act as an intelligent advisor, offering feedback in addition to some automated assistance. Synthesis tools can prompt programmers for additional information. However, how to best interact with scientists regarding different kinds of scientific programming tasks is an open question in need of further exploration.

\sidebar{An important relationship exists between program synthesis and explainable AI.}

An important relationship exists between program synthesis and explainable AI. State-of-the-art program synthesis techniques depend on deep learning and other data-driven approaches. As a result, the extent to which the program synthesis process itself is explainable depends on explainable AI, including machine learning techniques. Program synthesis itself, on the other hand, can  distill a data-driven learning and exploration process into a set of symbolic, understandable rules. These rules can form an explainable result from the machine learning process and thus serve as a technique for producing explainable AI processes.

\sidebar{Tools useful for scientific programming may need to leverage transfer learning techniques in order to apply knowledge from larger programming domains to scientific programming.}

\subsection{What are the challenges in applying program synthesis to scientific computing and HPC problems?}

\begin{itemize}
    \item \textbf{Small Sample Sizes: } Scientific programming comprises a small part of the overall programming market. This situation implies that tools useful for scientific programming, regardless of how the tool development is funded, may need to leverage transfer learning techniques in order to apply knowledge from larger programming domains to scientific programming. The challenges associated with the small sample sizes of scientific programs are exacerbated by the fact that scientific programs are effectively optimizing a variety of different objectives (e.g., performance, portability, numerical accuracy), and  how the relevant trade-offs were being made by the application developers often is unknown. The challenge of small sample size  applies not only to the code itself but also to the set of programmers; and since many synthesis tools are interactive, training data on how humans most effectively interact with the tools is required but challenging to obtain.
    \item \textbf{Lack of Relevant Requirements Languages: } There does not currently exist, either as an official or a de facto standard, an input language capable of expressing the requirements of a scientific application. While  work has been done on expressing correctness conditions in general-purpose software, this is often focused on correctness. Covering the performance requirements, numerical requirements, and so on needed for scientific applications has not been captured in a requirements language.  Work also has been done on extracting requirements from natural language, unit tests, and other informal artifacts associated with existing code, and these techniques should be applied to scientific applications as well; but in cases where a user might wish to directly supply requirements,  no broadly applicable way to do so currently exists. Significant domain knowledge is involved in constructing scientific applications, but this knowledge  is often not explicitly stated, such as how boundary conditions should be handled and how accurate the results need to be. This lack also is manifest in challenges getting different synthesis tools to interoperate effectively.
    \item \textbf{Verification and Validation: } Verifying that a scientific program satisfies its requirements is a complicated process, and validating the application is likewise a complicated but essential part of the scientific method. The lack of a suitable requirements language, the ambiguities inherent in natural language specifications, and other implicit parts of the code requirements make even defining what ``correct'' means a challenge. For many configurations of interest,  no analytic solution exists to which one can compare with certainty. Moreover, the use of randomized and nondeterministic algorithms, limited numerical stability, and the use of low-precision and approximate computing make even the process of comparing with a reference solution difficult. Scale also makes verification difficult: the computational resources required to test an application might not be readily or regularly acquired. Verification can take multiple forms, often all of which are important: {
    \begin{itemize}
        \item A mathematical verification (proof) of correctness
        \item The results of randomized testing showing no problems
        \item A human-understandable explanation of what the code does and why it is correct
    \end{itemize}
    } Verification often assumes that the base system functions correctly; but especially on leading-edge hardware, defects in the hardware and system software can be observed, and narrowing down problems actually caused by system defects is an important goal. How to measure comprehensibility, succinctness, and naturalness and otherwise ensure that code can be, and is, explained is an open question. For applications that simulate physical processes, the process of validating that the result matches reality sufficiently well can involve expensive and sometimes difficult-to-automate physical experiments.
    \item \textbf{Legacy Code: } Existing scientific code bases have large amounts of code in Fortran and C, generally considered legacy programming languages. Because of a lack of robust, modular components for processing code in these legacy languages and because of the smaller number of samples for training in these legacy languages, it can be challenging to apply program synthesis tools to code that must interact with these code bases. Moreover, even if modern languages such as C++ or Python are used, the libraries used for scientific development are often distinct from the libraries used more generally across domains. Some of these libraries use legacy interface styles (e.g., BLAS) that present some of the same challenges as do legacy programming languages.
    \item \textbf{Integration and Maintenance: } The use of tools that generate source code has long been problematic from an integration perspective, especially  if the code generation process is nondeterministic or depends on hardware-specific measurements or on human interaction. With all such tools, difficult questions must be addressed by the development team: {
      \begin{itemize}
          \item Should the tool be run as part of the build process?
          \item If run as part of the build process, builds become nondeterministic and perhaps slower, both because of the time consumed by the tool and because the requirements associated with making sufficiently reliable performance measurements may limit build parallelism.
          \item If not run as part of the build process, how is the output cached? Is it part of the source repository? How is the cache kept synchronized with the tool's input? Do all developers have access to all of the relevant hardware to update the cached output? Alternatively, can the tool generate performance-portable code? If the tool is interactive, how are invocation-to-invocation changes minimized?
      \end{itemize}
    }
    
    \sidebar{Synthesis tools face challenges, not only in interfacing with existing code, but also in interfacing with each other.}
    
    \item \textbf{Composability: } As with other software components and tools, synthesis tools face challenges, not only in interfacing with existing code, but also in interfacing with each other. The properties of code generated by one tool may need to be understood by another synthesis tool, and the tools may need to iterate together in order to find an overall solution to the programming challenge at hand~\cite{aleen:2009:asplos}.
    The Sparse Polyhedral Framework~\cite{Strout18} provides a possible foundation for composing data and schedule transformations with compile-time and runtime components in the context of program synthesis.
    
    \item \textbf{Stability and Support:} Like other parts of the development environment, program synthesis tools require a plan for stability and support in order to mitigate the risk that defects or missing features in the tools do not block future science work. In order to transition from research prototypes to tools useful by the larger scientific community, the tools must be built on robust, well-maintained infrastructures and must be robust and well maintained themselves.
    
    \item \textbf{Search Space Modeling for Program Synthesis:}
    The search space of program synthesis for scientific workloads is  complex, rendering many search algorithms ineffective. For example, in autotuning, the search can be formulated as a noisy, mixed-integer, nonlinear mathematical optimization problem over a decision space comprising continuous, discrete, and categorical parameters. For LLVM loop optimization, as an example, we are faced with a dynamic search space that changes based on the decisions at the previous step. 
    Modeling the search space of these problems will significantly reduce the complexity of the problem and will allow us to develop effective problem-specific search methods. Application- and architecture-specific knowledge should be incorporated as models and constraints. 
    Moreover, by considering metrics such as power and energy as objectives rather than constraints, we can obtain a hierarchy of solutions and quantify the sensitivities associated with changing constraint bounds. These can significantly simplify (or even trivialize) online optimization at runtime, when quantities such as resource availability and system state or health are known. 
    
\end{itemize}

\subsection{What are aggressive short-term, medium-term, and long-term opportunities and goals?}

A number of opportunities exist to apply program synthesis techniques to scientific-computing problems, and these opportunities will expand with time.

\subsubsection{Short Term (1--2 Years)}
\begin{itemize}
    
    \item \textbf{Defining Challenge and Benchmark Problems: } The community working to develop new program synthesis technologies can use challenge problems to direct their long-term aims and enable conversations with the scientific programming community. In order to measure  progress toward addressing those challenge problems, establish concrete examples, and enable comparison between systems, collections of benchmark problems should also be developed. Separate challenge and benchmark collections can be developed for different classes of problems (e.g., programming-language translation, specification-driven synthesis, and example-driven synthesis).
    
    \item \textbf{Interactive Synthesis, Repair, and Debugging: } Constrained techniques, such as proposing simple hole fillings \cite{DBLP:journals/pacmpl/LubinCOC20,An.AugmentedExampleSynthesis.POPL20} and local source-code edits to correct user-identified mismatches between expected and observed application behaviors, can be intcorporated into integrated development environment software that is being used to develop scientific software.
    
    \item \textbf{Smarter Code Templates: } Synthesis tools can provide assistance with generating boilerplate code, applying design patterns, and performing other largely repetitive programming tasks. Some of these patterns occur more frequently in scientific computing than in other domains, such as halo exchange in physical simulations.
    
    \item \textbf{Numerical Precision, Accuracy, and Stability: } Synthesis tools can start addressing situations in which different options exist between algorithmic variants and the precision used to represent numerical quantities. Automatic selection between these options, based on user-supplied metrics, including performance, accuracy, and stability, should be possible. A fundamental requirement is to guard synthesis via rigorous and scalable roundoff error estimation methods~\cite{das2020satire}.
    
    \item \textbf{Superoptimization and High-Quality Compilation: } Synthesis systems can use exhaustive search techniques effectively in restricted domains to find the best algorithmic compositions and code generation for particular systems. Superoptimization in restricted domains can apply to library call sequences and other high-level interface generation tasks. Compilation improvements will include more helpful and more accurate user feedback regarding where and what code annotations will be helpful.
    
    \item \textbf{Knowledge Extraction: } Analysis and verification tools will extract ever more precise and relevant high-level specifications from existing implementations, often referred to as lifting, and use these specifications to enable verification, interface generation, and other tasks~\cite{iyer:2020:PSG}.
        
    \item \textbf{Intelligent Searching: } Intelligent code search engines can be constructed that account for code structure, naming, comments, associated documents (e.g., academic papers), and other metadata. Finding code examples and other existing solutions to related problems can significantly increase programmer productivity.
    
    \item \textbf{Synthesis via Component Assembly:} The large-scale application frameworks in the past two decades have used componentization and assembly for a modest level of high-level program construction. The key idea is to have an infrastructural framework that becomes a backbone that model and algorithmic components of the solution system can plug into as needed. While this methodology by itself would not suffice for the multilevel parallelism that we are facing now, the concept of assembly from components can still provide an easy way of attacking some of the performance portability challenges by incorporating variable granularities in how componentization occurs. Past frameworks assumed components at the level of separate standalone capabilities in a quest to shield the science and core numerical algorithms from the details of infrastructure and assembly. Allowing componentization within such standalone capabilities (in a sense letting some of the infrastructural aspects intrude into the science sections of the code) can be helpful in reducing replicated code for different platforms. For example, one can view any function as having a declaration block, blocks of arithmetic expressions, blocks of logic and control, and blocks of explicit data movements that can each become a component. A subset of these components can have multiple implementations if needed or, better still, be synthesized from higher-level expressions. A tool that does not care whether the implementations  exist as alternatives or are synthesized will be relatively simple and general purpose and can have a huge impact on productivity and maintainability, while at the same time reducing the complexity burden on other synthesis tools in the toolchain.
    \item \textbf{Benchmarks:} Developing a set of easy-to-use benchmarks and well-defined metrics for comparison will be critical for advancing the algorithms for program generation. These benchmarks should reduce or hide the overhead required to let researchers from other areas to develop algorithms. For example, autotuning can benefit greatly from applied math and optimization researchers, but currently  no easy-to-use framework  allows these researchers to test algorithms.    
    
\end{itemize}

\subsubsection{Medium Term (3--5 Years)}
\begin{itemize}

    \item \textbf{Synthesis of Test Cases: } Based on user-specified priorities, background domain knowledge, and source analysis, synthesis tools can generate test cases for an application and its subcomponents. These span the granularity space from unit tests through whole-application integration tests and, moreover, can include various kinds of randomized testing (e.g., ``fuzz testing'') and the generation of specific, fixed tests.
    
    \item \textbf{Parallelization and Programming Models: } Synthesis tools can assist with  converting serial code to parallel code and converting between different parallel-programming models. Parallelism exists at multiple levels, including distributed-memory parallelism (e.g., expressed using MPI) and intranode parallelism (e.g., expressed using OpenMP, SYCL, or Kokkos).
    
    \item \textbf{Optimized Coupling Code: } Complex applications often require different subcomponents to communicate efficiently, and the ``glue code'' needed between different components is often tedious to write. Synthesis tools can create this kind of code automatically and over the medium term can create customized bindings that limit unnecessary copying and format conversions. Over the longer term, these customized data structure choices can permeate and be optimized over the entire application. 
    
    \item \textbf{Performance Portability: } Generating code that works well on a particular target architecture is challenging, but generating code that performs well across many different target architectures adds an additional layer of complexity. Synthesis should be able to address this combined challenge of generating performance-portable code (using, e.g., OpenMP, SYCL, or Kokkos) that has good performance across a wide array of different platforms.

    \item \textbf{Autotuning:}
    Autotuning is becoming a proven technology to achieve high performance and performance portability.
    Despite several promising results, however,  a number of challenges remain that we need to overcome for a wider adoption \cite{balaprakash2018autotuning}. Autotuning should be made  seamless and easy to use from the application developer's perspective. This task involves a wide range of advancements from  automated kernel extraction and large-scale search algorithms, to reducing the computationally expensive nature of the autotuning process. Modeling objectives such as runtime, power, and energy as functions of application and platform characteristics will play a central role. These models will be used to quantify meaningful differences across the decision space and to offer a convenient mechanism for exposing near-optimal spots in the decision space.

\end{itemize}

\subsubsection{Long Term}
\begin{itemize}
    
    \item \textbf{Solving Challenge Problems: } Challenge problems should be solvable by using composable, widely available tools. These tools should be capable of incorporating background knowledge from a wide variety of domains and of producing efficient, verifiable solutions in a reasonable amount of time. Where appropriate, the code will use sensible identifier names and otherwise be readable and maintainable.
    
    \item \textbf{Intentional Programming: } Synthesis tools can operate using high-level mathematical and natural language specifications, largely automatic but eliciting key feedback from human scientists, working within a common framework that supports the tooling ecosystem.
    
    \item \textbf{Lifelong Learning: } Synthesis tools will use an iterative refinement scheme, learning from user feedback and automatically improving themselves as time goes on. Synthesis tools will be able to create new abstractions for different domains and evolve them over time. Recent developments in reinforcement learning can be an effective vessel for lifelong learning.
    
    \item \textbf{Understanding Legacy Code: } Lifting, or the extraction of high-level features from concrete implementations, can work over large bodies of code. Synthesis tools will use lifting processes to understand legacy code, including any necessary use of background knowledge, and can interface with it or translate it to other forms.
    
    \item \textbf{Full-Application Verification: } Full applications, including library dependencies, can be verified, symbolically, statistically, or otherwise, with high confidence. The verification procedures can be driven automatically based on user-provided priorities and intelligent source-code analysis.
    
    \item \textbf{Proxy Generation:} Proxy applications, representing user-specified aspects of a full application, can be automatically generated by synthesis tools. These proxies will include appropriate source-code comments and documentation.

    \item \textbf{End-to-End Automation for Autotuning:}
    Autotuning needs to be part of the compiler tool                                                     chain. The process of autotuning should not involve any manual intervention. We need to develop autonomous, intelligent autotuning technology that can surpass human capability to accelerate computational science on extreme-scale computing platforms.

\end{itemize}

%% file: compiler_technology.tex
\section{Compiler Technology}

Synthesis technology and compiler technology are intimately tied. In the end, compilers are an essential consumer of the output of synthesis tools. Moreover, synthesis tools often reuse programming language processing and analysis infrastructure from compiler infrastructures in order to process inputs and mine existing code bases for patterns and other background knowledge. Improvements in synthesis technology are expected to go hand in hand with improvements in compiler technology.

\sidebar{Improvements in synthesis technology are expected to go hand in hand with improvements in compiler technology.}

\subsection{What opportunities exist for improving compilers and program analysis tools to better support scientific applications and HPC?}

Compilers can better leverage machine learning techniques to drive heuristics and other algorithmic tuning parameters. Examples include the following:
\begin{itemize}
    \item Optimizing the order in which transformations are applied
    \item Optimizing the thresholds and other parameters used by transformation and analysis routines, including choices between algorithms, to balance cost vs.\ benefit tradeoffs.
    \item Optimizing the sets of features used to drive the aforementioned tuning decisions.
    \item Building cheap surrogate models for performance modeling in autotuning, which  can be used to prune the large search space and identify promising regions in a short time. 
\end{itemize}

Compilers and other analysis tools can produce more information, both about the code being compiled and about the tool's modeling results and decisions. This information can be used by human engineers and by synthesis tools as part of an iterative development process.

As hardware architectures become more complex, with multiple levels in memory and storage hierarchies and different computational accelerators, compilers can offer additional support for mapping applications into the underlying hardware. This may require more information regarding data sizes, task dependencies, and so on than what is traditionally provided to a compiler for a programming language such as C++ or Fortran. Data sizes, for example, affect how code is optimized: the optimal code structure for data that fits into the L1 cache may be different from that for data that fits into the L2 cache. These code structure differences include loop structures, data structures and layouts, and accelerator targeting.

\subsection{How might higher-level information be leveraged to capture those opportunities?}

A significant challenge in implementing compilers for Fortran, C++, and other high-performance languages used in scientific computing is the lack of higher-level information in the source code. The compiler generally does not know the size of data arrays, the likely location of data within the memory subsystem, and the number of threads being used. This lack of information forces compiler developers to fall back on heuristics that are likely to work across a larger number of use cases instead of creating models that can optimize specific use cases.

When DSLs are used, some of this information is contained in the input, and where it is not, the semantics of the DSL can constrain the number of specialized versions that the compiler might generate to a practical number. Similarly, known library interfaces can effectively form a DSL and have similar properties that can be extracted by an intelligent compiler to leverage in driving the compilation process. Separate compilation, the general case where an application's source code is split across multiple separately compiled source files, is helpful in enabling parallelism in the build process and other separations of concerns but limits the higher-level information that can be usefully extracted. Tools that build application-scoped databases from source-analysis processes can potentially mitigate that loss of information. These tools, along with associated source annotations and known library interfaces, can enable high-level compiler optimizations while remaining transparent to the programmer.

\subsection{How might compiler technology be improved to better integrate with program synthesis systems?}

Synthesis systems are generally iterative, combining techniques to search the space of potential solutions with techniques for evaluating particular candidate solutions. Compiler technology can be enhanced for better participation in both parts of this process. Information extracted from code compilation or attempted compilation of partial solutions can be used to constrain the search. In addition, information from compilation can be used to evaluate potential solutions  in terms of both correctness and performance; however, performance models from the compiler can be used to inform a wider set of metrics.

Recent advances in compiler frameworks, such as LLVM and its multilevel IR (MLIR), can be leveraged to extract suitable information from applications. Examples of features that can be easily extracted are properties of loop-based computations that exhibit regularity, such as loop depth, array access patterns, shape and size of multidimensional arrays, and dependence structures (e.g., dependence polyhedra in affine programs). As these features evolve and change within and among intermediate representations of a compiler,  metadata must be collected describing the sequence of transformations applied. Similarly, for more irregular application such as those in arising in sparse linear systems, the sparsity pattern and structure can be used as synthesis features and can be either  collected during the system assembly process or provided by the end user via compiler directives.

Integrating with an iterative synthesis process may also require new schemes to be implemented in compilers that allow for the caching of partial compilation and analysis results so that each variant evaluated during the synthesis process does not incur the full overhead of performing the necessary analysis and transformations each time.

\sidebar{When both a synthesis tool and the underlying compiler are responsible for code optimization, it is an open question how this responsibility is best divided and how the cooperation will be arranged.}

When both a synthesis tool and the underlying compiler are responsible for code optimization, it is an open question how this responsibility is best divided and how the cooperation will be arranged. Synthesis systems might be responsible for higher-level transformations that are difficult to prove correct in the absence of higher-level information, for example, or for the use of hand-tuned code fragments that a machine-programming system is unlikely to discover automatically. As work continues, open interfaces that are adopted by multiple compilers and synthesis tools will likely enable a diverse, vibrant ecosystem of programming-environment capabilities.